\documentclass[runningheads]{llncs}
\usepackage{graphicx}

\usepackage{tikz}
\usepackage{comment}
\usepackage{amsmath,amssymb} 
\usepackage{color}
\usepackage{mathtools}
\usepackage{bbm}

\usepackage[width=122mm,left=12mm,paperwidth=146mm,height=193mm,top=12mm,paperheight=217mm]{geometry}

\begin{document}
\pagestyle{headings}
\mainmatter

\title{Semi-TCL: Semi-Supervised Track Contrastive Representation Learning\\} 


\author{
Wei Li\and
Yuanjun Xiong \and
Shuo Yang\and
Mingze Xu\and
Yongxin Wang \and
Wei Xia
}

\institute{{\tt\small wayl, yuanjx, shuoy, xumingze, yongxinw, wxia@amazon.com}\\Amazon AWS AI}

\maketitle
\begin{abstract}
Online tracking of multiple objects in videos requires strong capacity of modeling and matching object appearances. 
Previous methods for learning appearance embedding mostly rely on instance-level matching without considering the temporal continuity provided by videos. 
We design a new instance-to-track matching objective to learn appearance embedding that compares a candidate detection to the embedding of the tracks persisted in the tracker.
It enables us to learn not only from videos labeled with complete tracks, but also unlabeled or partially labeled videos.
We implement this learning objective in a unified form following the spirit of constrastive loss.
Experiments on multiple object tracking datasets demonstrate that our method can effectively learning discriminative appearance embeddings in a semi-supervised fashion and outperform state of the art methods on representative benchmarks.

\end{abstract}

\section{Introduction}
Online multiple object tracking (MOT) usually performs three tasks simultaneously: a) object detection; b) motion prediction; c) appearance matching (also known as Re-Identification (ReID)). 
Previous methods implement these three functions either separately, such as the earlier works using different off-the-shelf models~\cite{wojke2017simple}, or in an integrated way. For example, recent works on combining motion prediction~\cite{bergmann2019tracking,stone2000centertrack} or appearance modeling~\cite{zhang2020fairmot} as additional heads on an object detection backbone.  
Among these methods, obtaining representative appearance features is a central topic. 

The appearance representation is used for matching a newly detected object instance to a set of objects being tracked at a certain time-step. The appearance module needs to have strong discriminative power to distinguish the ``same'' object from other objects despite the inter-instance and intra-instance variations. Earlier approach~\cite{wojke2017simple} utilizes separately trained ReID~\cite{zheng2016mars} models for this purpose. Recently, Zhang \textit{et al}~\cite{zhang2020fairmot} propose to learn the appearance embedding using a classification task and demonstrated that this integrated model can achieve good tracking performance. Nonetheless, the existing methods to learn appearance embedding mostly draw inspiration from image-level instance recognition tasks, such as face recognition~\cite{ouyang2015deepid,deng2019arcface} or ReID~\cite{li2017learning}.
That is, the learning objective is usually to match one object instance, in the form of an encoded image patch, to another instance in the same object track (metric learning~\cite{chen2020simple,sohn2016improved,hoffer2015deep}), or its corresponding ``class'' indexed on the object's identity. These methods are limited in several aspects. 
First, the instance-to-instance matching objective does not utilize the temporal continuity of video.
This is because such method stems from image-level recognition datasets where the temporal information is not present. 
Second, existing appearance embedding learning methods require complete track annotations for training, which is laborious to obtain for a sufficient amount of videos. 
These issues call for a method that can 1) utilize the temporal information in videos for learning appearance representation and 2) learn from both labeled and unlabeled videos.

We present a Semi-supervised Track Contrastive embedding learning approach, Semi-TCL., a new method for learning appearance embedding to address the above issues. 
We start by devising a new learning objective of matching a detected object instance to a track formed by tracked object instances in a few video frames. 
This design fits closely to the actual use case of appearance embedding where each newly detected instance will be matched against the aggregated representation of tracks. 
It also alleviates the need for full track-level annotation of videos for learning. 
Low-cost primitive trackers can be used to generate track labels on unlabeled videos, which can be used together with fully annotated but scarcely labelled videos. 
We show that effectively learning using the instance-to-track objective can be implemented with a form of contrastive loss~\cite{khosla2020supervised}, where the tracks serve as the positive samples and negative samples for contrasting. This unified loss formulation can be applied to all videos regardless of whether they are fully annotated, achieving practical semi-supervised learning. 
Semi-TCL can be applied to state-of-the-art online MOT models with integrated detection and appearance modeling, which provides a simple end-to-end solution for training MOT models.

We benchmark tracking models learned with Semi-TCL on multiple MOT datasets, including MOT15~\cite{leal2015motchallenge}, MOT16~\cite{milan2016mot16}, MOT17~\cite{milan2016mot16}, and MOT20~\cite{dendorfer2020mot20}. Our model outperforms other state-of-the-art tracking methods on all benchmarks. 
We further study the effect of several design choices in Semi-TCL and demonstrate that it can effectively learn from unlabeled videos through semi-supervised learning and the proposed instance-to-track matching objective is more suitable for learning appearance models for the MOT task.

\begin{figure}[t]
      \centering
      \includegraphics[width=.75\linewidth]{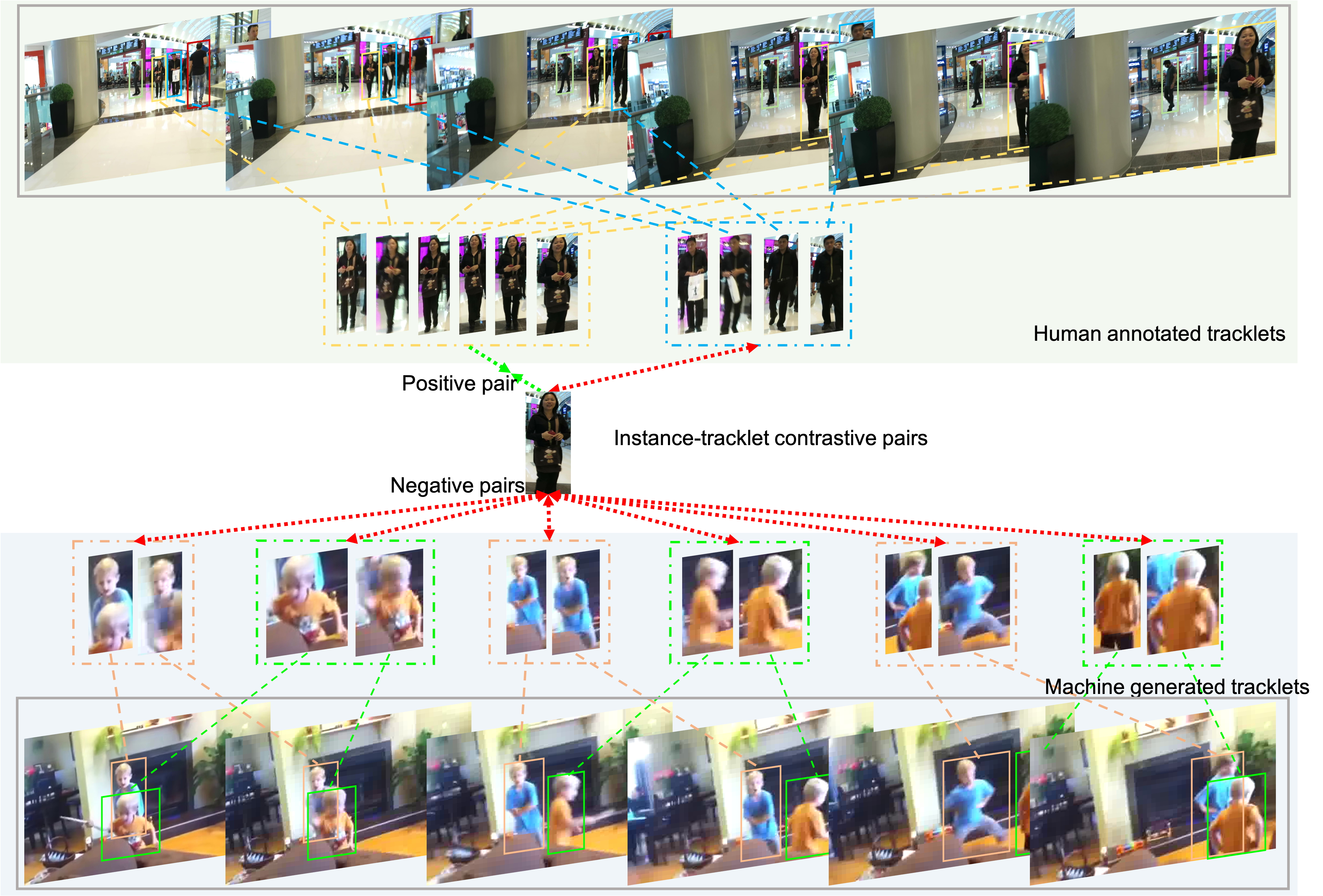}
      \centering
      \caption{Illustration of the instance to track matching learning objective.(better view in color). Instance (person in the middle) matches with both labeled tracks (top) and unlabeled tracks (bottom, label computed by model prediction), with this objective, distance of positive pairs (green line) will be pushed together and negative pairs (red line) are pulled apart. }
      \label{fig:teaser}
\end{figure}

\section{Related Work}

\textbf{MOT and ReID.}
With the rapid development of deep learning, we keep witnessing new breakthroughs in MOT areas. Wojke~\textit{et al.}~\cite{wojke2017simple} employed the tracking by detection idea and provided a fast online tracking method. \cite{bergmann2019tracking,stone2000centertrack} combined the detection and tracking module and proposed the joint detection and tracking approach. These approaches provide multiple options to generate the tracklets. To connect the tracklets, ReID embedding learning is a necessary component.
\cite{li2014deepreid} explored the detection and ReID embedding work, while the detection and ReID learning are separate so it is not efficient. \cite{xu2018joint,shuai2020multi,wang2019towards,zhang2020fairmot} jointly detect and learn the ReID embedding, improving the overall runtime significantly. Currently, joint learning of multiple object tracking and ReID tend to be the most efficient solution, we follow this design in our work. However, different from these works which rely on complete human labeled tracking data, we conduct a semi-supervised learning manner.

\textbf{Contrastive embedding learning.}
Contrastive learning \cite{khosla2020supervised,sun2019learning,khosla2020supervised,tian2020makes,he2020momentum,hadsell2006dimensionality,wu2018unsupervised} had been studied for a long time for the visual embedding learning. Researchers used to build a local batch and construct positive pairs from the same class and negative pairs for the different ones. They try to push apart negative embedding distances and squeeze the positive ones. \cite{khosla2020supervised} proposed a loss for supervised learning while still building on the self-supervised contrastive methods by leveraging the label information. To build the positive pairs, \cite{khosla2020supervised} looked into the instances in a batch and construct positive and negative pairs based on class labels. SCL~\cite{khosla2020supervised} unified the real labeled data and unlabeled data in one format. SCL allows both supervised and unsupervised learning to follow the same formation and permits jointly combining the labeled data and partially labeled data learning together. This makes \cite{khosla2020supervised} outperform the baseline where cross entropy is used in the image classification tasks. MoCo\cite{he2020momentum} is another important contrastive learning approach which focuses on building a dynamic dictionary to boost the contrastive learning. Our work is inspired by the flexibility of dealing with image labels proposed by\cite{khosla2020supervised}. We employed the contrastive idea and proposed a unified objective, which is shared by both labeled and unlabeled video in ReID embedding learning.

\textbf{Video/Group embedding learning.}
Video embedding learning is widely investigated in video related tasks. \cite{qian2020spatiotemporal} proposed a video contrastive learning approach leveraging spatial and temporal cues to learn spatial-temporal representations from unlabeled videos. \cite{sun2019deep} proposed a self-supervised learning approach for video features. The work proved that the learned features are effective for a variety of downstream video tasks, such as classification, captioning and segmentation. Video based ReID learning has also been investigated. \cite{chen2018video} proposed the competitive snippet-similarity aggregation and temporal co-attentive embedding. With the design, intra-person appearance variation is reduced and similarity estimation is improved by utilizing more relevant features. Yang \textit{et al} \cite{yang2020spatial} proposed a Spatial and Temporal Graph Convolution Network to learn the ReID embedding from video sequence. By jointly extracting structural information of a human body and mining discriminative cues from adjacent frames, the approach achieved state-of-the-art results on Video ReID benchmarks \cite{zheng2016mars,wu2018cvpr_oneshot}. \cite{lan2020semi} proposed a semi-online approach to tracking multiple people. The method employed Multi-Label Markov Random Field and focused on efficiently solving the ReID learning in challenging cases. The video based embedding shows that the temporal information from video is helpful in learning embeddings. As we are trying to learn embedding from tracking videos, employing temporal information from sequence might be beneficial.

    

\section{Method}

For the task of online MOT, earlier methods~\cite{wojke2017simple} usually utilize separately learned visual embedding model from either person~\cite{prosser2010person,li2014deepreid,yi2014deep} or face recognition~\cite{liu2016large,deng2019arcface,wang2018cosface} tasks. The models are trained mostly on image datasets, which may suffer from the large domain gap between image and video data. 
Recent works started to investigate joint learning of the visual feature for ReID together with other components in an integrated tracking model~\cite{zhang2020fairmot}.
We aim at building models that simultaneously perform object detection and tracking using appearance features. Similar to \cite{zhang2020fairmot}, we build our joint model on top of CenterNet \cite{duan2019centernet}. 
An ID branch with two convolution layers operates in parallel to the heatmap prediction branch in~\cite{duan2019centernet} to perform visual feature extraction at each center location. The visual feature is extracted from the detection centers for matching newly detected object instances to objects being tracked by the tracker.

The overall loss function for training our model is 
\begin{align}
    \boldsymbol{L_{joint}} = \boldsymbol{L_{det}} + \boldsymbol{L_{id}},
\end{align}
where $\boldsymbol{L_{det}}$ is the loss for the object detection branch and $\boldsymbol{L_{id}}$ denotes the loss for visual embedding learning. 
We use the same loss formulation from \cite{duan2019centernet} for $\boldsymbol{L_{det}}$ on every video frame in training and design a novel way of constructing $ \boldsymbol{L_{id}}$ and learning visual embedding.


\subsection{Learning with Instance-to-Track Matching}

Existing separate and joint visual embedding learning methods mostly start from an image-level instance matching problem. 
That is, they try to learn an embedding function $f(\cdot)$ that maps each image $I$ to a $C$-dimensional vector $f(I)$ with a certain distance metric, which is usually the $\ell_2$ distance. 
Given two images or image crops depicting the appearance of two object instances, $I_1$ and $I_2$, 
we expect $f(I_1)$ and $f(I_2)$ to have a small distance when they are showing the same object and otherwise a large distance.
Traditionally, learning the embedding function is achieved by comparing each image to other images of the same or different object. 
One can use either a classification loss
\begin{align}
    \mathcal{L}_{class} = - \log y_i\widehat{p}_{y_i},
\label{eq:class}
\end{align}
or the margin loss~\cite{liu2016large,elsayed2018large}
\begin{align}
\mathcal{L}_{margin} = \mathbbm{1}_{y_1 = y_2}\|f(I_i) - f(I_j)\|_2^2 + \mathbbm{1}_{y_1 \ne y_2}\max(0, m -\|f(I_i) - f(I_j)\|_2^2).
\label{eq:margin_contrast}
\end{align}
Here $y_i$ denotes an instance's identity label and $\widehat{p}_{y_i}$ is the classification output probability of $I_i$ being to identity class $y_i$ out of the $K$ identity classes.
For example in~\cite{zhang2020fairmot}, Eq.~\ref{eq:class} is used to classify one detected instance to $K$ potential classes, the annotations of which are obtain by labeling all tracks in all videos across all training datasets. 

Now consider the case of using the learned visual embedding in online tracking. At each time-step $t$, a newly detected object instance needs to be matched to a set of existing tracks. 
But each track $\mathbf{T}_j$ usually contains multiple instances of the tracked object accumulated over time. An additional aggregation function $G(\cdot)$ has to be introduced to make this matching possible. 
Thus the matching is actually between $f(I_i)$ and the aggregate track-level $G(I_j^{t_0}, \ldots, I_j^{t-1})$,
where $I_j^t$ denotes the instance of the object depicted by track $T_j$ at time $t$.
The added aggregation function is apparently not addressed in the original learning objective of image level matching, as in Eq.~\ref{eq:class} or Eq.~\ref{eq:margin_contrast}.
Thus, using the visual embedding learned by either one for the matching in online track could be sub-optimal. 

To address this discrepancy, our learning objective should be directly built on the aforementioned instance-to-track matching task. 
Formally, for a temporally ordered set of object instances $T_j = \{I_j^0, \ldots, I_j^n\}$ that belong to the same object $j$, we defined the aggregation function $G(\cdot)$ that maps the set of features $\{f(I_j^0), \ldots, f(I_j^n)\}$ to a single vector $\mathbf{g}_j$.
We learn the embedding function $f(\cdot)$ and the aggregation function $G(\cdot)$ so that the object to track distance 
\begin{align}
    D(I_i, T_j) = d(f(I_i), G(f(I_j^0), \ldots, f(I_j^n))). = d(f(I_i), \mathbf{g}_j)
\end{align}
is small when $I_i$ and $T_j$ are depicting the same object and large otherwise. 
Explicitly incorporating the aggregation function into the learning objective has two advantages: 1) it makes the learning objective close to the actual tracking scenario, which enables the embedding learning to benefit from the temporal information in videos; 2) as we shall see later, it make it easier to extend the learning objective to videos with partial or without track level annotations. 

\subsection{Tracklet Contrastive Learning}
Given one object instance $I_i$, there is a track $T_i$ that this instance belongs to. 
$T_i$ contains multiple instances $G(I_i^0, \ldots, I_i^l)$, where $l$ is the length of the track. 
We can generate random sub-tracks $S(i) = \{\Tilde{T}_i^j\}$ of $T_i$ by sampling random subsets of instances $T_i$ . 
These sub-tracks resemble the actual partial tracks occur in online tracking. That is,
at a given time-step during online tracking, we can only observe a portion of the complete track that have already been shown in the video. 
For a batch of input videos $V$, we can sample a set of object instances  $\mathbf{I} = \{I_0, \ldots, I_N\}$ belonging to their corresponding sub-tracks $\mathbf{\tilde{T}} =\{\tilde{T}_0, \ldots, \tilde{T}_L\}$.
With these $N$ instances and $L$ subtracks, we can implement the instance-to-track matching  objective in the contrastive loss form
\begin{align}
    \mathcal{L}_{TCL} = \sum_{i=1}^N \frac{-1}{|S(i)|}\sum_{\tilde{T}_j \in S(i)} \frac{\exp{(f(I_i)\cdot \mathbf{\tilde{g}}_j / \tau)}}{\sum_{\tilde{T}_l\in \mathbf{\tilde{T}}} \exp{(f(I_i)\cdot \mathbf{\tilde{g}}_l / \tau)}}.
\label{eq:i2tloss}
\end{align}
Here $S(i)$ denotes all sub-tracks that are sampled from the tracks that $I_i$ belongs to.  $\mathbf{\tilde{g}}_j$ is the aggregated visual feature of a sub-track $\tilde{T}_j$. We assume the feature vectors are all $\ell_2$normalized and the temperature parameter $\tau$ controls the scaling of the cosine similarities between vectors. We use $\tau=0.07$ following the general practices for contrastive losses. 

We call the proposed method of learning visual features in a tracker with Eq.~\ref{eq:i2tloss} as tracklet contrastive learning (TCL). Compared with instance-level contrastive learning~\cite{wu2018unsupervised,khosla2020supervised,he2020momentum} which compares one image to another image, the instance to track loss has two different concepts in the comparison: the object instances and the sub-tracks. 
Because this type of comparison is close to the actual use case in tracking, we expect the learned visual features to be more suitable to the ReID task during online tracking. 
In this work, we use a simple aggregation function $G$ that averages all input feature vectors, which we empirically found to give satisfying visual embeddings. 
But TCL does not inhibit the uses of more advanced aggregation functions which could be developed in future.

\subsection{Learning with Labeled and Unlabeled Videos}\label{learn_with_videos}
Learning with the instance-to-track matching objective also enables us to extend the learning task to videos without human annotated track labels. 
In Eq.~\ref{eq:i2tloss}, we notice that only the sampled sub-tracks, instead of the complete tracks, are used in training. 
On the other hand, when we apply a certain primitive multiple object tracker that relies on motion prediction to videos without track-level annotations, we can obtain a large amount of potentially incomplete tracks.
The generation of these incomplete tracks can be viewed as sampling sub-tracks from the complete, annotated tracks, which have not been annotated. 
This means the seemingly unusable unlabelled videos have now become a potential source for mining useful sub-tracks in TCL.
In particular, for videos with no track annotation, we can apply a motion prediction based tracker~\cite{stone2000centertrack} and obtain a set of predicted tracks. 
These tracks are treated as the pseudo labels of these videos. We can then train our tracker using these pseudo-labeled videos together with the annotated videos. 

\begin{figure}
  \begin{center}
    \includegraphics[width=0.55\textwidth]{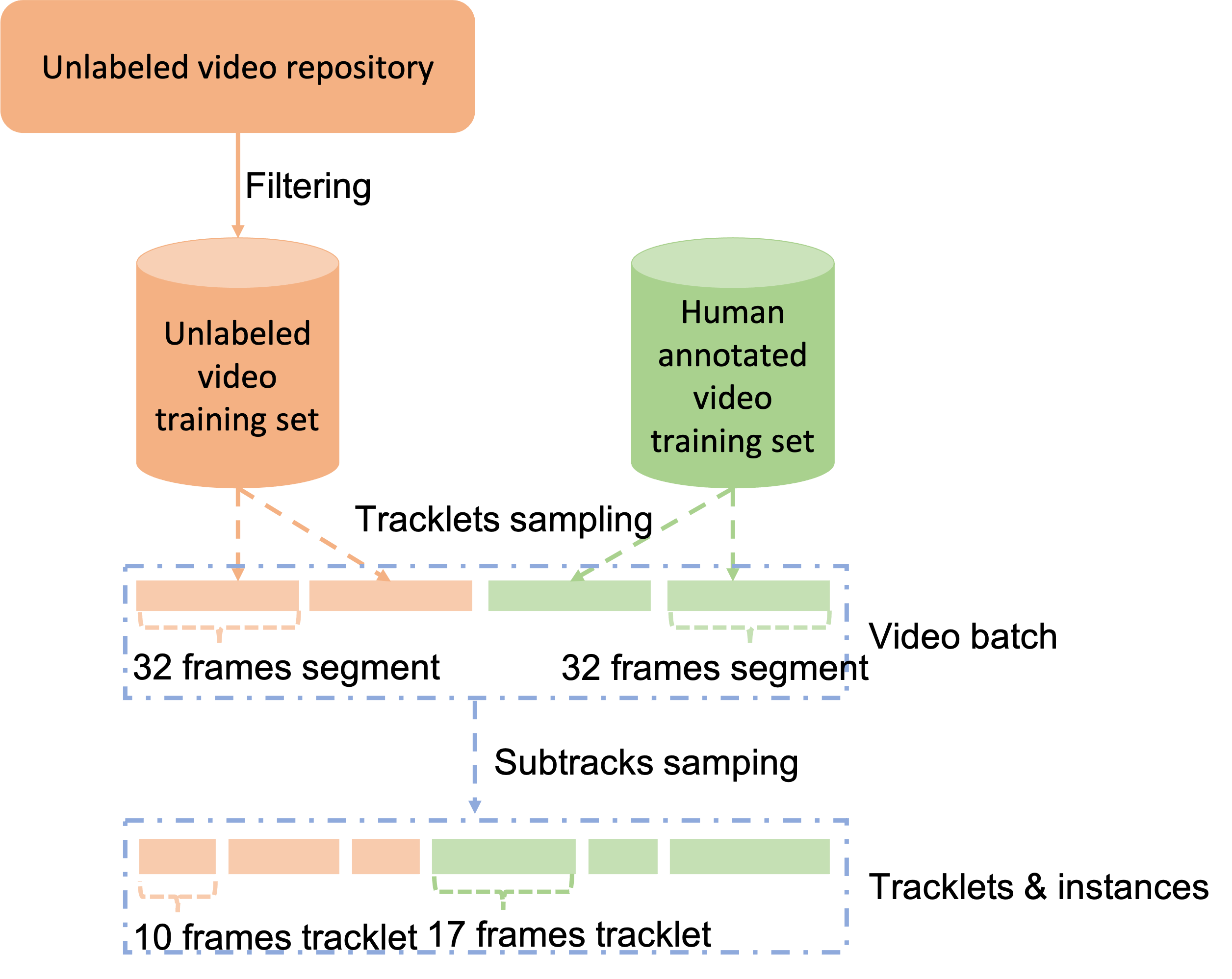}
  \end{center}
  \caption{Training instance and tracklets sampling strategy. We filter the unlabeled videos by the number of tracks predicted. They are then used towgether with the labeled videos in Semi-TCL. Each training batch contains 4 video segment of 32 frames each with a $1:1$ ratio between labeled and unlabeled videos.}
  \label{fig:training_strategy}
\end{figure}

Formally, we obtain a track annotated video set $\mathbf{V}_A$ and an unlabeled video set $\mathbf{V}_U$ for learning with Semi-TCL.
Usually the unlabeled video set is much larger than the labeled set but may contain segments that has very few objects of interest, thus has less value for learning.
We apply a primitive tracker such as~\cite{stone2000centertrack} on $\mathbf{V}_U$ to obtain predicted tracks for each video. 
Then we rank the unlabeled videos in $\mathbf{V}_U$ by the number of produced tracks in them.
To mine potentially useful videos for Semi-TCL, we simply take the top-$K$ videos based on tracklet density in the rank and produce a refined video set $\mathbf{V}_R$.
This $\mathbf{V}_R$ is used together with  $\mathbf{V}_L$  in training. 
We split each video in both $\mathbf{V}_A$ and $\mathbf{V}_R$ into segments of $32$ consecutive frames. 
We randomly sample $2$ segments from  $\mathbf{V}_A$ and $2$ segments from $\mathbf{V}_R$ in each training step to form one training mini-batch. 
From these $4$ segments, we can obtain $M$ tracks, either annotated or produced by the primitive tracker. 
We perform another round of sampling on these tracks so that for each track we can obtain $3$ sub-tracks, meaning $L=3M$. 
This ensures that each instance is exposed to multiple sub-tracks of the same track. 
We extract $N$ object instances from these sampled sub-tracks. 
These samples are then used for calculating the loss in Eq.~\ref{eq:i2tloss}.
This process is illustrated in Fig.~\ref{fig:training_strategy}.
The loss function in Eq.~\ref{eq:i2tloss} is differentiable and easy to empirically optimize. Thus models with Semi-TCL can be learned with backprogagation in an end-to-end manner.

\section{Experiments}
\subsection{Dataset and metrics} 



In the Semi-TCL experiments, three types of dataset are used. They are image detection dataset for pre-training, labeled video tracking dataset for supervised joint tracking and embedding learning, unlabeled video dataset for Semi-supervised learning.

\noindent\textbf{Person detection dataset} we employed Crowdhuman\cite{shao2018crowdhuman} for the pre-training. Crowdhuman is a person detection image dataset with more than 20k images and 470K instances. 

\noindent\textbf{Labeled video tracking dataset} We used MOT15, MOT17, MOT20 training set as our labeled set. MOT15, 17, 20 are from multiple academic datasets and annotated with human tracking information. The dataset are widely used for supervised tracking and ReID. 

\noindent\textbf{Unlabeled video dataset} We employed the AVA-Kinetics\cite{li2020ava} and MEVA \cite{corona2021meva} Dataset to boost the Semi-TCL learning. MEVA and AVA-Kinetics are originally used for human activity detection. The AVA-Kinetics dataset has relatively low resolution varying from $144\times256$ to $454\times256$, and the total amount of videos is 230k. 
 We select 3 sets of videos from AVA-Kinetics with the amount of 100, 200, 300 based on the tracklet density. The total frame number for the three selected sets are 24755, 49135, 73923 respectively. Compared with AVA-Kinetics, MEVA dataset has a high resolution with $1920\times1072$. We select 15 of the videos with a total number of 17754 frames for training.

We report IDF1, MOTA, MT, ML, IDS on MOT series test benchmarks. Among the metrics, we prioritize IDF1 and MOTA over other metrics as it corresponds closely with the embedding learning. On the test benchmarks, we report our results on the private session based on our results obtained from the MOT challenge server. On our ablation studies, we report IDF1, MOTA and IDS to compare the impact of different components.

\begin{table}[]
\caption{Evaluating Semi-TCL the private detection tracks of the MOT challenge benchmarks. We report results evaluated by the public evaluation servers. Bold fonts mark the best results. }
\centering
\begin{tabular}{ |p{2cm}||p{1cm} p{1cm} p{1cm} p{1cm} p{1cm} p{1cm}|  }

 \hline
 \multicolumn{7}{|c|}{MOT15 test} \\
 \hline
 Methods & IDF1 & MOTA & IDS & MT & ML & Frag \\
 \hline
 FairMOT\cite{zhang2020fairmot} & 64.7 & 60.6 & 591 & \textbf{343} & 79 & 1731\\
 GSDT\cite{wang2020joint}   & 64.6 & \textbf{60.7}  & \textbf{477} &339 & 76 & 1705\\  
 TubeTK\cite{pang2020tubetk}   & 53.1 & 58.4 & 854 & 283 & 130 & \textbf{1194}\\
 Semi-TCL & \textbf{64.9} & 60.6 & 551 & 344 & 88&  1687 \\
 \hline
 \multicolumn{7}{|c|}{MOT16 test} \\
 \hline
 Methods & IDF1 &MOTA & IDS &MT & ML & Frag\\
 \hline

 DeepSort\cite{wojke2017simple}   & 62.2 & 61.4 & \textbf{781} & \textbf{249} & 138 & 2008  \\
 TubeTK\cite{pang2020tubetk}   & 59.4 & 64.0 & 1117 & 254 & 147 & \textbf{1366} \\
 CTracker\cite{peng2020chained} & 57.2 & 67.6 & 1897 & 250 & 175 &  3112\\
 GSDT\cite{wang2020joint} & 69.2 & 66.7 & 959 & 293 &144 & 2596 \\
 FairMOT\cite{zhang2020fairmot}& 72.8 & \textbf{74.9} & 815 & 306 &127 & 2399\\
 Semi-TCL & \textbf{73.9} & 74.8 & 925 & 322 & \textbf{130} & 2569 \\
 \hline
  \multicolumn{7}{|c|}{MOT17 test} \\
 \hline
 Methods & IDF1 &MOTA & IDS &MT & ML & Frag\\
  \hline
 
 SST\cite{sun2019deep}   & 49.5 & 52.4 & 8431 & \textbf{504} & 723 & 14797\\
 TubeTK \cite{pang2020tubetk}   & 58.6 & 63.0 & 4137 & 735 & 468 & \textbf{5727}\\
 Ctr.Track \cite{stone2000centertrack} & 64.7 & 67.8 & 3039 & 816 & 579 & 6102\\
 CTracker \cite{peng2020chained} & 57.4 & 66.6 & 5529 & 759 &570 & 9114\\
 GSDT \cite{peng2020chained} & 66.5 & 73.2 & 3891 & 981 &411 & 8604\\
 FairMOT\cite{zhang2020fairmot} & 72.3 & \textbf{73.7} & 3303 & 1017 & \textbf{408} & 8073\\
 Semi-TCL & \textbf{73.3} & 73.3 & \textbf{2790} & 972 & 441 & 8010 \\
 \hline
   \multicolumn{7}{|c|}{MOT20 test} \\
 \hline
 Methods & IDF1 &MOTA & IDS &MT & ML & Frag\\
  \hline
 FairMOT\cite{zhang2020fairmot}& 67.3 & 61.8 & 5243 & 855 &\textbf{94} & \textbf{7874}\\
 GSDT \cite{wang2020joint}& 67.5 & \textbf{67.1} & \textbf{3131} &\textbf{660} &164 & 9875\\
 Semi-TCL & \textbf{70.1} & 65.2 & 4139 & 761 & 131 & 8508 \\
 \hline
\end{tabular}
\label{tab:main_eval_table}
\end{table}

\subsection{Implementation details}\label{sec:implementation}
A 8 NVIDIA Tesla V-100 GPUs machine is used to train the Semi-TCL models. We select 144 as our batch size and the starting learning rate is 1e-3. Person detection dataset is first employed as pre-training, and then Semi-TCL training is conducted on the joint set of labeled and unlabeled videos. We train the Semi-TCL model for 200 epochs before dropping learning rate to 1e-4, and for another 20 epochs until the training fully converges. 
For the unlabeled video, we use Center Track \cite{stone2000centertrack} preprocessed 20k videos from AVA-Kinetics and 15 MEVA videos, tracking threshold is set to be 0.3 to process all the videos. From the 20k processed AVA videos, we select 100/200/300 videos based on a tracklet density based mining strategy. To make sure unlabeled data not dominate the training, we applied a balanced sampling strategy based on the method in \ref{learn_with_videos}.

\subsection{Comparison with State of the Art}
Semi-TCL is trained on the joint labeled and unlabeled video dataset, while tested on MOT15, MOT16, MOT17, MOT20 benchmarks. With the MOT benchmarks test annotations unavailable, we submit our test prediction results to the MOT server and obtain our benchmarks results. Table \ref{tab:main_eval_table} shows the benchmark results of Semi-TCL as well as other SOTA approaches. Since our work focuses on ReID embedding learning for tracking, the primary metric for us is the IDF1. Based on Table \ref{tab:main_eval_table}, our method consistently outperforms the other the state of the art approaches in all MOT benchmarks. Specially, on MOT16 and MOT17, Semi-TCL is able to have 1\% and 1.1\% increase under the IDF1 metric. On MOT20 where the dataset tends to have very crowded scenes and ReID is highly relied to match trackletd, our method improves the SOTA IDF1 score from 67.5\% to 70.1\% with a 2.5\% improvement. It is also worth noticing that, in all four MOT benchmarks, we have the best score in IDF1, which highlights the quality of the ReID embedding. The comparison of the test results with other SOTA approaches shows the superiority and robustness of Semi-TCL.

\subsection{Design choices in TCL}
As the core component of this work, TCL based a instance to tracklet matching scheme instead of the widely used instance to instance matching during contrastive pair building. To show the effectiveness of the work, we start the ablation study with the comparison of TCL with other instance matching based approach. All the comparison experiments are using half of MOT17 as labeled tracking training data and the other half for validation.

\begin{table}[tb]
        \caption{MOT17 validation ablation study on loss objective and batch size. CE refers to the Cross Entropy pretraining on detection dataset. SCL means the instance-to-instance match based contrastive embedding learning. CE pre and SCL pre stands for the CE and SCL applied to detection dataset based pretraining. TCL represents the tracklet contrastive embedding learning. TCL w. b32 means the TCL learning with the batch size to be 32, same applied to b96 and b144.} \label{tab:loss_comparison}
        \begin{tabular}{llll}
 & IDF1 & MOTA  & IDS  \\ \hline
 CE pre     & 48.2   &  47.0 & 463       \\
 SCL pre    & 53.6   &  45.2    &   404         \\ 
\hline

CE & 74.7    & 70.5 & 404  \\
SCL & 75.5 & 74.7 & 365 \\
TCL & 76.2   & 74.6 & 339\\
\hline

TCL w. b144 & 76.2   & 74.6 & 339\\
TCL w. b96  & 75.1   & 73.1 & 358\\
TCL w. b32 & 74.4 & 70.4 & 321\\
\end{tabular}
      \centering
        \caption{MOT17 validation evaluation on external datasets. TCL stands for tracklet contrastive learning with only labeled MOT17 training data. TCL w.AVA100 stands for Semi-TCL training with the joint labeled MOT17 training set and unlabeled AVA100 external dataset. Same applied to TCL w.AVA200 and TCL w.AVA300.  TCL w.MEVA means Semi-TCL training with the joint labeled MOT17 training set and unlabeled MEVA dataset. And TCL w.AVA+MEVA means Semi-TCL training with the joint labeled MOT17 training set and combined AVA100 and MEVA.}\label{tab:semi_table}
        \begin{tabular}{llll}
 & IDF1 & MOTA  & IDS  \\ \hline

 TCL         & 76.2   &  74.6 & 339       \\
 \hline
 TCL w.AVA100      & 76.9   &  74.9    &   310         \\ 
 TCL w.AVA200      & 77.2   &  74.2    &   343         \\ 
 TCL w.AVA300      & 77.8   &  74.1    &   352         \\
\hline

TCL w.MEVA & 78.1    & 77.6 & 423  \\
TCL w.AVA+MEVA & 78.4    & 78.0 & 375  \\

\end{tabular}
\end{table}

\noindent\textbf{Contrastive loss vs. other instance recognition losses}. To see whether the proposed embedding learning objective is effective, we compare the performance of different embedding learning objectives. Our baseline method is the cross entropy (CE) objective function, which is common in many computer vision applications and proven to be effective for embedding learning. In tracking embedding learning, with the tracking labeled data, images from the same tracklet are regarded as samples of same class. We also compare with the a baseline contrastive learning objective using instance-to-instance match, referred to as SCL~\cite{khosla2020supervised}. They are compared with the TCL objective in Table \ref{tab:loss_comparison}.
We report comparison result on MOT17 validation set of the different objective functions. We can see that TCL outperforms both CE and SCL objective functions. We also notice that the MOTA are similar between SCL and TCL, but the IDF1 score improves from 75.5 to 76.2. It suggests that the instance to trackl matching objective could be more effective for the ReID learning.

\textbf{Impact of batch size on training}
Larger batch sizes tend to be useful in image embedding contrastive learning tasks\cite{chen2020improved}. We would like to see if this also holds in the scenario of tracking embedding learning. We use 3 batch sizes for comparison, 32, 96, 144. We keep the training setting same as the main experiment and only use batch size as variable. Evaluation results can be found in \ref{tab:loss_comparison}. We find that while increasing the batch size from 32 to 96 and 144, the MOTA and IDF1 have a 0.7\%, 2.7\% and 1.1\%,1.5\% improvement respectively. This means larger batch sizes, or more contrastive learning pairs, are helpful to the tracking embedding learning.

\subsection{Semi-supervised learning with TCL}

\textbf{Pre-training comparison} 
Static image pre-training is proved to be useful in the joint tracking and ReID learning \cite{zhang2020fairmot}. We also applied contrastive learning based pre-training in our approach. To see whether CE or contrastive learning can have better pre-training quality, we use Crowdhuman dataset \cite{shao2018crowdhuman} for training and evaluation is conducted on MOT17 validation. Table \ref{tab:loss_comparison} demonstrates the benchmark results, we can see the SCL based approach can outperform the CE based approach significantly in IDF1 with a 5.4\% gap. In MOTA the SCL is behind for 1.8\%, which shows that the SCL help learned better embedding quality. 


\noindent\textbf{Accuracy vs. volume of unlabeled videos} The effectiveness of Semi-TCL assumes that external video is helping the embedding learning. We want to move one step further, figuring out the relationship of Semi-TCL learning with different number of videos. 
Setting the total learning epoch to be 150, 200, 300 for AVA100/200/300 respectively, we can obtain three Semi-TCL models. Results can be see in Table \ref{tab:semi_table}. Not to our surprise, with more additional videos, we do see the improvements in the IDF1 from 76.9 to 77.8. It is interesting to observe that the MOTA does not have obvious change with more data, staying around 74. This is understandable as no additional human supervision is provided for the detection task. Table \ref{tab:semi_table} shows embedding learned with MEVA dataset (15 videos and 17k frames) outperform all the three AVA series dataset with much smaller data amount. With this comparison, we are also interested in the case where we combine the AVA and MEVA datasets. By combining the MEVA and AVA100 dataset, we found the joint video dataset can boost the MOT17 evaluation results further to IDF1 78.4 and MOTA to 78.0.
From the AVA and MEVA experiments we can see that the tracklet contrastive learning objective can benefit from the increasing number of unlabeled video data.

\begin{figure}[t]
      \centering
      \includegraphics[width=1\linewidth]{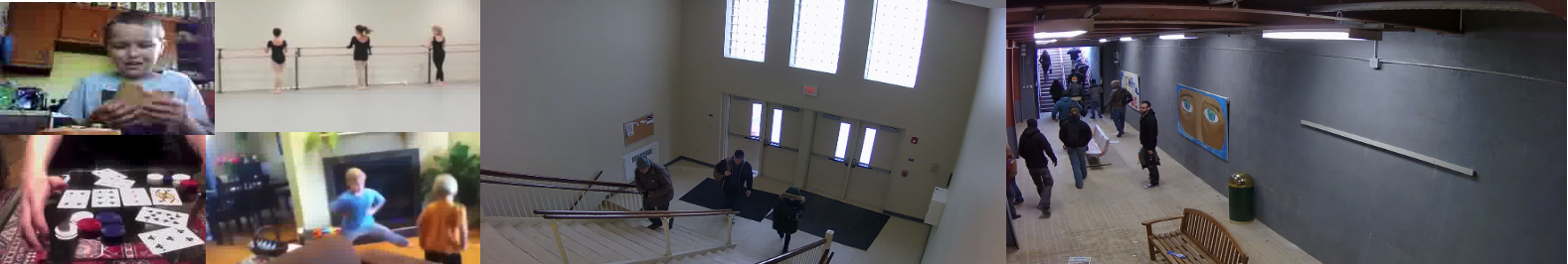}
      \centering
      \caption{Snapshots of images in AVA and MEVA (left two columns/four images are from AVA-Kinetics and right two from MEVA, resolution are $454\times256$ $1920\times1072$, the images also shows the MEVA video has more appearance and motion diversity).}
      \label{fig:dataset_cmp}
\end{figure}

\noindent\textbf{Accuracy vs. types of videos}. Besides the volume of videos, the unlabeled videos may come from different domains. MEVA \cite{corona2021meva} and AVA are both curated as action recognition datasets but the content type is different \ref{fig:dataset_cmp}. With larger resolution than AVA videos and crowded scenes, videos in the MEVA dataset are more akin to the videos presented in MOT dataset, where the video are mostly from surveillance or car mounted cameras.  
Comparing the results of semi-supervised learning with either of the two datasets, we observe that unlabeled videos with similar content are more effective in increasing the tracking accuracy. 

\noindent\textbf{Mining strategy for unlabeled videos}.
We sample the unlabeled video based on tracklet density as we believe more predicted tracklet might mean more human related content. Based on on primitive prediction results, the mining dataset has average 103 tracks while the overall average tracks number is 36.7. To verify if this tracklet density based video mining strategy help the embedding learning, we conduct a ablation study to compare dataset with mining v.s. random selection. To run this experiment, we also build a AVA 100/200/300 dataset by just random selection. We can observe training with filtered videos, which have more tracks produced by the primitive tracker, leads to better increase in accuracy with respect to the number of videos used. 

\begin{figure}
      \centering
      \includegraphics[width=\linewidth]{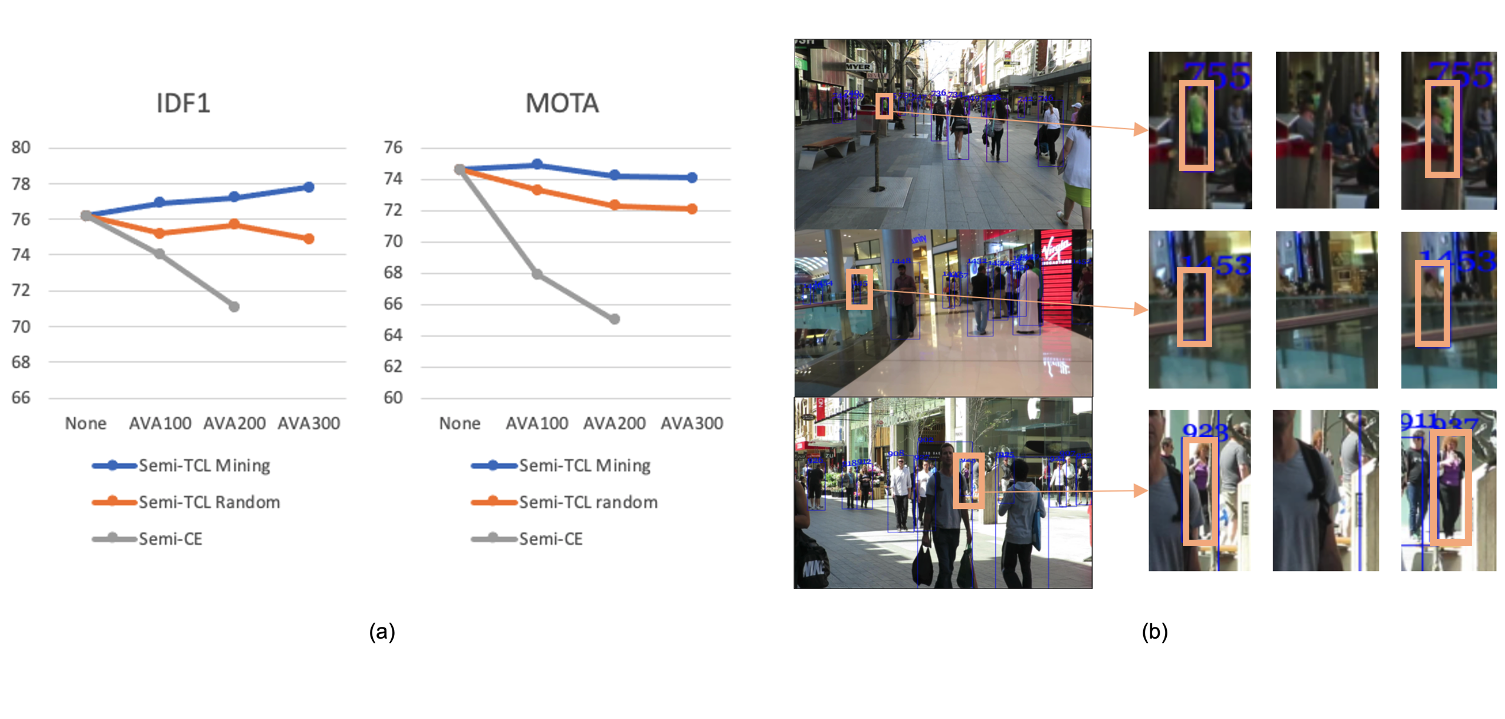}
      \centering
      \caption{(a)IDF1 and MOTA results with unlabeled data growth. We use blue, red, gray color to represent the metrics for Semi-TCL with video dataset with mining, Semi-TCL with random selection from videos and using CE on video set with mining. (b)Visualization of error examples in Semi-TCL predictions. In the left section, we show the starting frame and use bounding box highlight the regions the example happen. In the right section we zoom in and crop the region. Row 1 show Semi-TCL successfully match a new detection to the previous track ID with partial detection. Row 2 shows the model fails and match to a wrong ID due to blueness. Row 3 shows the rematch fails after long time when the missing track expires.}
      \label{fig:datasetgrowth_cmp}
\end{figure}

\noindent\textbf{Use of contrastive loss for semi-supervised learning}. We also compared Semi-TCL with an alternative approach which uses the cross entropy loss in~\cite{zhang2020fairmot} (CE) for semi supervised learning on the AVA and MOT17 training joint dataset. We show the IDF1 and MOTA results in Figure \ref{fig:datasetgrowth_cmp} to compare with the results learned via Semi-TCL. Both methods are trained with the mined unlabeled videos and labeled videos as decribed above. We can observe that CE seems to be not benefiting form additional unlabeled videos. So we stop adding more videos to it. In contrast, Semi-TCL continually benefits from more unlabeled videos.

\subsection{Error Analysis}\label{sec:limit}
We demonstrate qualitative results of the Semi-TCL on MOT test samples. In Figure \ref{fig:datasetgrowth_cmp}(b), we show a positive sample in the first row and two error samples in second and third row. In the first row, we find the person with track \#255 can be correctly re-identified after being occluded for one frame. In the second row, the region is extremely blurred which deteriorates the visual repsentation quality. As a result, Track \#1452 is first assigned to a person in black coat then matched with a person in yellow. Example in the third row shows a case where a person is occluded for a extended period of time and thus can not be correctly associated with his previous track. 
The error sampls shows though we have achieved good improvement in tracking accuracy, there still exist several challenging situations that remains to be tackled in future research works.

\section{Conclusion}
In the paper, we proposed Semi-supervised tracklet level embedding learning approach (Semi-TCL). Semi-TCL extends the embedding learning from instance-instance match to instance-tracklet match which fits more closely to how ReID embedding is used in tracking. Semi-TCL uses the contrastive loss to implement this idea and is able to learn embeddings from both labeled video and unlabeled videos. Evaluation of Semi-TCL on MOT15, MOT16, MOT17, MOT20 shows the state of the art performance on all the benchmarks, which is further justified by our ablation studies. We observe an promising growth of accuracy when the amount of unlabeled videos increases, which may shed light on large-scale semi-supervised or unsupervised learning of multiple obejct tracking models.

\clearpage
\bibliographystyle{plain}
\bibliography{arxiv_main}


\end{document}